\pdfoutput=1

\documentclass[11pt]{article}

\usepackage[preprint]{coling}

\usepackage{times}
\usepackage{latexsym}

\usepackage[T1]{fontenc}

\usepackage[utf8]{inputenc}

\usepackage{microtype}

\usepackage{inconsolata}

\usepackage{graphicx}

\usepackage{tabularx}
\usepackage{multirow}
\usepackage{subcaption}
\usepackage{array} 
\usepackage{algpseudocode}
\usepackage{amsmath}
\usepackage{wrapfig}
\usepackage[utf8]{inputenc} 
\usepackage[T1]{fontenc}    
\usepackage{hyperref}       
\usepackage{url}            
\usepackage{booktabs}       
\usepackage{amsfonts}       
\usepackage{nicefrac}       
\usepackage{microtype}      
\usepackage{xcolor}         
\usepackage{enumitem}
\usepackage{listings}
\usepackage{colortbl}
\usepackage{pifont}

\title{Look, Compare, Decide: Alleviating Hallucination in Large Vision-Language Models via Multi-View Multi-Path Reasoning}

\author{Xiaoye Qu$^{1*}$, Jiashuo Sun$^{2}$\thanks{Equal contribution.}, Wei Wei$^1$\thanks{Corresponding author.},  \textbf{Yu Cheng}$^3$   \\  
 $^{1}$ Huazhong University of Science and Technology 
 $^{2}$ Xiamen University \\ $^{3}$The Chinese University of Hong Kong \\
 \texttt{\{xiaoye, weiw\}@hust.edu.cn}; 
\texttt{gasolsun36@gmail.com}; \\ 
\texttt{chengyu@cse.cuhk.edu.hk} 
 }

\begin{document}
\maketitle
\begin{abstract}
Recently, Large Vision-Language Models (LVLMs) have demonstrated impressive capabilities in multi-modal context comprehension. 
However, they still suffer from hallucination problems referring to generating inconsistent outputs with the image content. To mitigate hallucinations, previous studies mainly focus on retraining LVLMs with custom datasets.
Although effective, they inherently come with additional computational costs. 
In this paper, we propose a training-free framework, \textbf{MVP}, that aims to reduce hallucinations by making the most of the innate capabilities of the LVLMs via \textbf{M}ulti-\textbf{V}iew Multi-\textbf{P}ath Reasoning. 
Specifically, 
we first devise a multi-view information-seeking strategy to thoroughly perceive the comprehensive information in the image, which enriches the general global information captured by the original vision encoder in LVLMs.  
Furthermore, during the answer decoding, we observe that the occurrence of hallucinations has a strong correlation with the certainty of the answer tokens. 
Thus, we propose multi-path reasoning for each information view to quantify and aggregate the certainty scores for each potential answer among multiple decoding paths and finally decide the output answer. 
By fully grasping the information in the image and carefully considering the certainty of the potential answers when decoding, 
our MVP can effectively reduce hallucinations in LVLMs.
The extensive experiments verify that our proposed MVP significantly mitigates the hallucination problem across four well-known LVLMs. 
The source code is available at: \url{https://github.com/GasolSun36/MVP}.
\end{abstract}

\begin{figure}[!t]
\setlength{\belowcaptionskip}{-1.5em}
\centering
\includegraphics[width=0.40\textwidth]{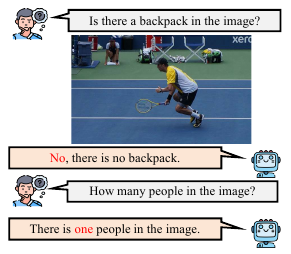}
\caption{Given an image, LVLM fails to recognize objects or miscounts the quantity.} 
\label{Figure-query}
\end{figure}

\section{Introduction}

Large Vision-Language Models (LVLMs) 
have become indispensable and marked a significant milestone in the field of Artificial Intelligence. 
These LVLM models, owing to their ability to generate contextually relevant textual renditions of visual inputs, are being extensively employed across a diverse spectrum of applications, such as healthcare~\cite{liu2023qilin,li2024llava,bazi2023vision}, autonomous systems~\cite{cui2024survey,tian2024drivevlm,park2024vlaad}, and robotics~\cite{liu2024online,shah2023lm,kelly2024visiongpt}.

Despite substantial advancements, LVLMs suffer from a significant challenge termed “hallucination”, whereby the models produce semantically plausible but factually inaccurate text, misaligned with the ground-truth content of the associated image.
As shown in Figure 1, LVLMs fail to recognize ``backpack'' and incorrectly identify the number of people in the image.
In applications where precision and reliability of generated content are paramount, such hallucinations can trigger a cascade of erroneous decisions. 
Consequently, addressing the hallucination issue is indispensable for strengthening the trustworthiness of LVLMs across practical applications.

To tackle hallucination, 
most recent studies focusing on retraining the LVLMs with constructed hallucination-related datasets by supervised fine-tuning (SFT) \cite{chen2023mitigating,wang2024mitigating,park2024mitigating,liu2023mitigating}, or Reinforcement Learning from Human Feedback (RLHF) \cite{yu2023rlhf,yan2024vigor,sun2023aligning}.  
Although these methods for alleviating hallucination in LVLMs have shown effectiveness, 
they acquire a substantial number of high-quality examples for training and are quite time-consuming and labor-intensive.
{Recently, there are also works exploring training-free paradigms to mitigate the hallucinations. Woodpecker \cite{yin2023woodpecker} pick out and correct hallucinations from the generated text. 
MARINE \cite{zhao2024mitigating} employs classifier-free guidance to incorporate the additional object grounding features to improve the precision of LVLMs’ generations. 
However, most of them heavily rely on external complicated tools, such as Grounding DINO \cite{liu2023grounding}, or BLIP-2-FlanT5X \cite{li2023blip}.}

In this work, to alleviate hallucinations in LVLMs, we focus on maximizing the innate ability of LVLMs without introducing additional training costs or external tools. 
To this end, we propose a novel training-free framework MVP, namely \textbf{M}ulti-\textbf{V}iew Multi-\textbf{P}ath Reasoning. 
Different from previous works, our MVP is grounded in an analysis of the key factors underlying hallucination, including the incomplete comprehension of image content and low certainty when decoding answer tokens in original LVLMs. 
First, if the vision encoder of LVLMs can not fully capture the information from the input image, language models may generate outputs based on this incomplete content, thus resulting in hallucinatory descriptions. 
Second, during the answer decoding, hallucinations occur more frequently when the certainty of answer tokens is low. In this scenario, the model is uncertain about multiple candidate tokens, leading to potentially inaccurate outputs. 

Thus, our MVP proposes to fully capture the information in the image and carefully consider the certainty of the potential answers when decoding. 
Specifically,
we first devise a multi-view information-seeking strategy, which involves an exhaustive perception of the image from varying dimensions: 
a "top-down" look captures overarching scene context, a "regular" view addresses elementary visual information, and a "bottom-up" perspective zooms in on intricate details. 
Instead of using tools, the captured information from these diversified views is generated by the LVLMs, and effectively reinforces the global image context captured by the original vision encoder of LVLMs, thereby reducing the hallucinations from misunderstanding the image's information.
In addition, during the answer decoding stage, 
we further introduce multi-path reasoning for each information view by explicitly quantifying the certainty score of the potential answers and then aggregating the overall certainty among multiple paths. 
Then, the answer with the highest certainty score will be chosen as the final answer, thus effectively alleviating the hallucinations caused by low certainty. 
To verify the effectiveness of MVP, we conduct experiments on four widely-used LVLMs. 
The promising results demonstrate that our framework significantly outperforms recent training-free methods. 

To sum up, our contributions are summarized as:

\begin{itemize}
    \item 
    We propose a training-free framework to alleviate hallucinations with Multi-view Multi-path Reasoning. Our framework focuses on maximizing the innate ability of LVLMs without introducing additional training costs or external tools. 
    
    \item To comprehensively grasp the image, we seek information from multi-view perspectives, including ``bottom-up'', ``regular'', and ``top-down'' views. During decoding, we introduce multi-path reasoning to quantify and compare the certainty of each potential answer.

    \item Through comprehensive experiments, we demonstrate the superior performance of our MVP in alleviating hallucinations across four LVLMs. Moreover, our framework is plug-and-play and can integrate with other decoding methods for further improvement.
\end{itemize}


\begin{figure*}
\centering
\includegraphics[width=0.96\textwidth]{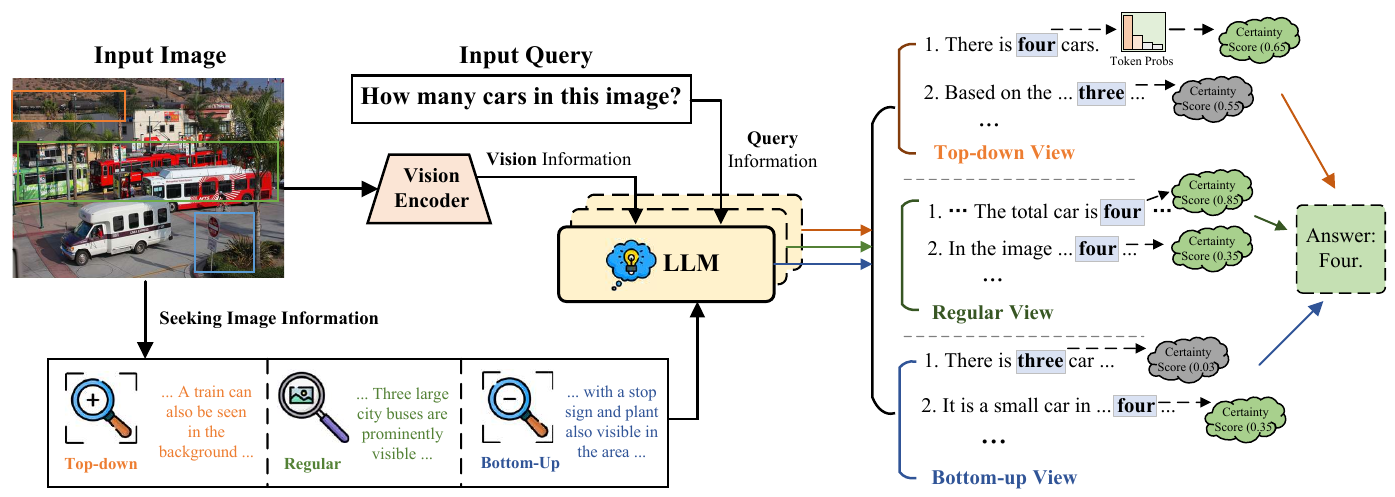}
\caption{An overview of our Multi-View Multi-Path Reasoning. (1) Seeking image information from multiple perspectives including top-down, regular, and bottom-up views.  
(2) Augmenting the global vision information with each view information.
(3) The certainty-driven decoding corresponding to each view quantifies and aggregates certainty scores for each potential answer among multiple decoding paths. The final results are obtained by comparing certainty scores among all candidates.}
\label{fig:framework}
\end{figure*}

\section{Method}

\subsection{Overall of the MVP Framework}

As shown in Figure 2, 
given that hallucinations commonly arise due to incomplete comprehension of image content, 
we propose to seek complementary information from the input image with three different views. 
Subsequently, the acquired information is leveraged to augment the global vision information from the vision encoder for LLM reasoning. 
For each view, considering different decoding paths have different certainty for potential answers, we introduce certainty-driven multi-path reasoning, which quantifies and aggregates the certainty score for each potential answer among multiple decoding paths. In this stage, we maximize the inherent reasoning ability of the model.
Finally, with the multi-view information and multi-path reasoning, we achieve superior performance for alleviating hallucinations.

\subsection{LVLMs Input and Decoding}

The input of LVLMs contains both image and text.  
The image is first processed by a vision encoder (e.g. CLIP \cite{radford2021learning}, BLIP \cite{li2022blip}) to obtain visual tokens. 
Then, the image tokens are mapped to the input space of LLMs for decoding. 
We denote the visual tokens as $\mathbf{x}^v = \{x_1^v,x_2^v,\ldots,x_{N}^v\}$. Here $N$ is the length of the visual tokens.  
Correspondingly, the input query is tokenized with the tokenizer. We denote it as $\mathbf{x}^q = \{x_1^q,x_2^q,\ldots,x_{M}^q\}$ with length $M$.
The image and text tokens are concatenated as the final input sequence $X$ with length ${N+M}$. 

\begin{equation}
    X = [x^v:x^q] = [{x_1^v, x_2^v,\ldots,x_{N}^v, x_1^q,x_2^q,\ldots,x_{M}^q}],
\end{equation}

After feeding the input tokens $X$ to the LVLMs, the model outputs answers in an auto-regressive manner which predicts the next token based on previous tokens, formally:

\begin{equation}
    p(O_t|O_{<t}) = \text{SoftMax}[LVLM([\{O_i\}_{i=1}^{t-1})],
\end{equation}
where we omit the input query $X$ and $\{O_i\}_{i=1}^{t-1}$ are decoding tokens from the previous $t$-1 rounds and the first decoding token $O_1$ is decoded with the input $X$ in Eq. 1. 
At time step $t$, the token with the highest probability is chosen from the vocabulary. During the decoding period, hallucinations arise when probabilities are improperly attributed to tokens that fail to correlate with the presented visual image.


\begin{figure}
\centering
\includegraphics[width=0.40\textwidth]{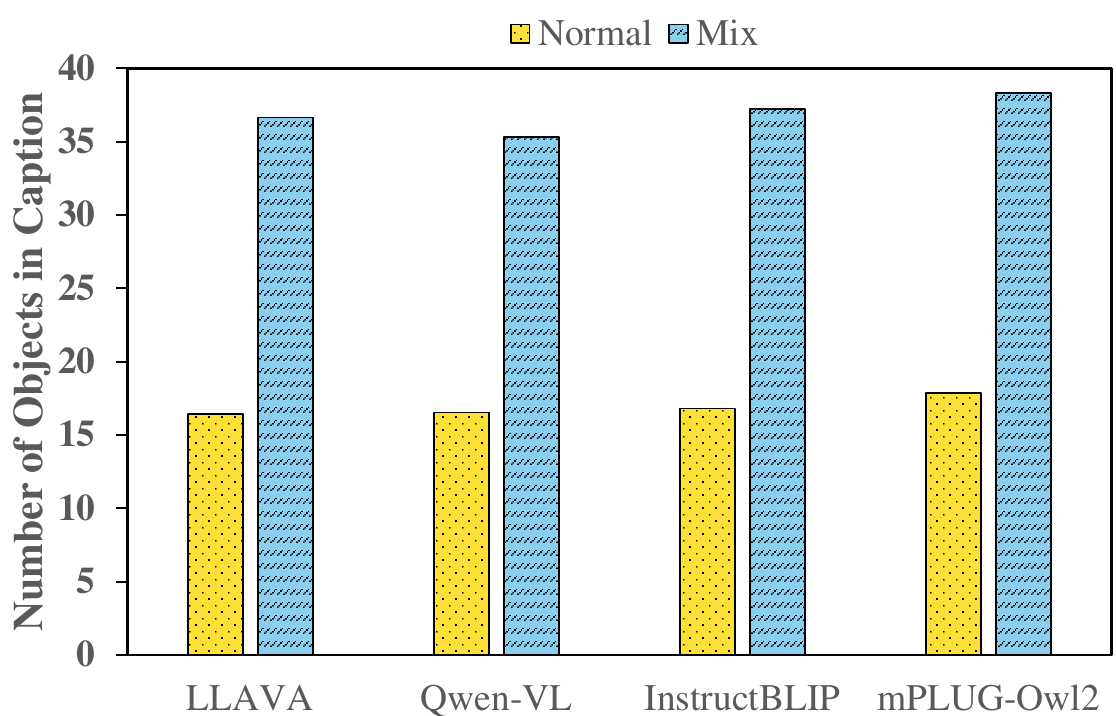}
\caption{Comparison of the number of objects between regular and multi-view caption. 
The statistic is obtained in MSCOCO Popular part of POPE benchmark.}  
\label{fig: object_caption}
\end{figure}


\subsection{Multi-view Image Information Seeking} 

Previous LVLM research, utilizing a CLIP for global image representation, may neglect intricate, object-specific details and background components, consequently leading to hallucinations precipitated by a partial grasp of the input imagery \cite{zhang2024avoiding}.
For instance, when querying the detailed information that is not captured by the video encoder, the LVLMs tend to hallucinate. 
Thus, it is imperative to master comprehensive information about the image before responding to the input query. Naturally, a wealth of visual information exists in images and can be located by various methods, such as invoking external visual detection tools \cite{liu2023grounding,he2017mask}. 

In this paper, we resort to maximizing using the innate ability of the LVLMs  
and design three perspectives for extracting comprehensive information: ``bottom-up'', ``regular'', and ``top-down''. 
To accomplish it, we use the given LVLM to generate captions by designing dedicated prompts, thus eliminating the need for external tools or the design of specific networks. For example, to extract the information from a top-down perspective, we use the prompt: ``\textit{Given the overall scene depicted in the image, taking into account the context, environmental factors, and any relevant visual cues, describe this image in details.}'' {(Please see Section 4.3 for more prompts)}.
To demonstrate the effectiveness of multi-view information-seeking strategy, we conduct a statistical analysis of the visual richness of multi-view captions. 
As shown in Figure \ref{fig: object_caption}, the LLaVA-1.5 model captures an average of 16.43 objects per image using only regular perspective caption, while an average of 36.66 objects can be recognized when three perspectives are adopted. 
Formally, the captions from a specific perspective can be tokenized and denoted as 
$\mathbf{x}^c = \{x_1^c,x_2^c,\ldots,x_{K}^c\}$ with length $K$ and $c \in \{ \text{Top-down}, \text{Bottom-up}, \text{Regular}\}$. Subsequently, the caption is integrated with the input for LLM decoding:


\begin{equation}
    X' = [{x_1^v,\ldots,x_{N}^v,
    x_1^c,\ldots,x_{K}^c,
    x_1^q,\ldots,x_{M}^q}],
\end{equation}


\subsection{Multi-path Certainty-driven Reasoning}

Decoding strategies are important in guiding how LVLMs produce textual answer. Previous decoding strategies commonly consider each output token with the same level of importance, thus ignoring the unique importance of the answer token.
However, we observe that the answer tokens present different certainty during diverse decoding paths. 
In Figure \ref{fig:framework}, for the question (How many cars are in this image), the first decoding path of ``Bottom-up'' and ``Regular'' perspectives produce different answers ``four'' and ``three'' but their certainty is significantly different (0.65 and 0.03, respectively). This phenomenon indicates that hallucinations occur more frequently when the certainty of answer tokens is low and 
inspires us with certainty-driven reasoning to alleviate hallucinations. 
Formally, we quantify the \textit{Centainty Score $S$}, which is the difference between the probabilities of the two tokens with the highest probabilities at time step $t$:

\begin{equation}
S = p(x^1_t \mid v, x_{<t}) - p(x^2_t \mid v, x_{<t})
\end{equation}

where $x^1_t$ and $x^2_t$ represent the post-softmax probabilities of top-two 
tokens at each decoding step $t$. It is worth noting that we only consider the probability disparity of the answer tokens.

\begin{figure}[!t]
\setlength{\belowcaptionskip}{-1.5em}
\centering
\includegraphics[width=8.5cm]{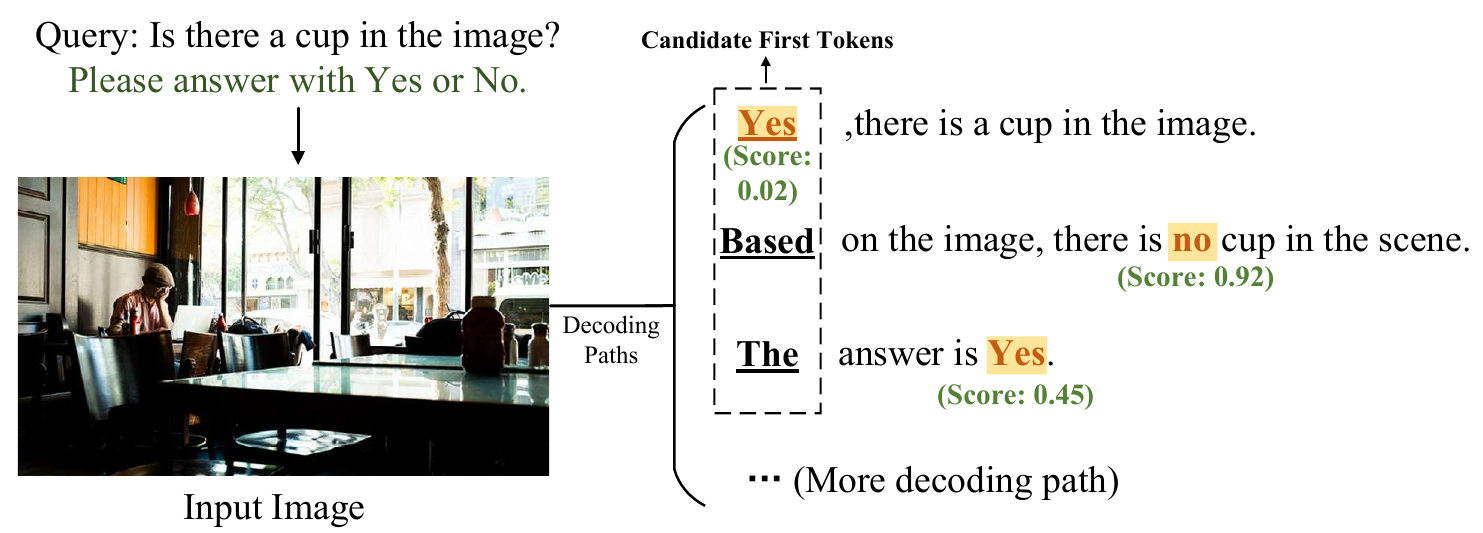}
\caption{
An illustration of certainty-driven multi-path reasoning. The correct answer is ``No". ``Score" denotes the certainty score of the answer token. ``Yes'', `Based'', ``The'' are candidate decoding tokens at first place. The three decoding paths are greedy decoding with these candidate tokens.
}  
\label{Figure-query} 
\end{figure}

\subsubsection{Multi-Path Certainty-driven Reasoning}

To illustrate certainty-driven reasoning, we first consider a basic situation where only one greedy decoding paths exist.
As shown in Figure 5, 
given the input query,
we observe that LVLMs tend to hallucinate when a low certainty score occurs, where greedy decoding mistakenly takes the bottle for a cup and outputs the wrong answer: ``Yes, there is a cup in the image", while the certainty score of the answer token ``Yes" is only 0.02.
With further investigation, when decoding the first token, besides ``Yes", there are many other candidates (i.e. ``Based", ``The"), which are displayed by underline in Figure 5 and sorted by probability from high to low. 
Instead of introducing complex methods to building multiple decoding paths, we simply inspect more top-$K$ paths starting from relatively lower probability tokens, namely decoding from the second word ``Based", the third word ``The", and so on. Notably, the second path leads to the correct answer ``no" with a significantly higher certainty score of 0.92.

Thus, we introduce a multi-path reasoning which explicitly considers the certainty of the answer tokens. Specifically, to build multiple paths, we consider the top-$K$ candidates in the decoding process of the first token, and then continue decoding based on each candidate to generate the $K$ paths with different answers. 
Formally, each path corresponds to an answer $A_{k}$. Here the answers can be identified by the question type or specified prompt format. For instance, we can search for numbers in the output to answer the question in Figure 2, or identify ``yes” or ``no" in Figure 5. Then, 
we aggregate the certainty score for same $\hat{A}_{k}$ from $K$ paths: 
\begin{equation}
S_{\hat{A}_{k}} = \sum^{K}_{j=1} ~~M_{j}(p(A_{{j}}^1 \mid v, A_{<t}) - p(A_{{j}}^2 \mid v, A_{<t}))
\end{equation}

where $A_{<t}$ denotes the sequence before generating answer $A_j$. Here $M_j$ is an optional parameter denoting the probability of the first token in the $K$-th path and we will explore it in the experiment. Thus, we can obtain the certainty score for each potential answer.

\subsubsection{Multi-View Multi-Path Reasoning} In Section 3.3, we seek image information from three perspectives $c \in \{ \text{Top-down}, \text{Bottom-up}, \\ \text{Regular}\}$. Considering that each view captures different information from the input image, the corresponding reasoning paths also present specific preference, thus we can further aggregate certainty scores for multi-view multi-path:

\begin{equation}
\small
S_{\hat{A}_{c, k}} = \sum^{c}_{i=1}\alpha_{i}\sum^{K}_{j=1} ~~ M_{ij}(p(A_{x_{ij}}^1 \mid v, A_{<t}) - p(A_{x_{ij}}^2 \mid v, A_{<t}))
\end{equation}

where $\alpha_{i}$ is a hyperparameter denoting the importance of a specific perspective.
Finally, the answer with the highest certainty score is selected as our final answer:
\begin{equation}
A_{final} = argmax(S_{\hat{A}_{c, k}})
\end{equation}

\begin{table*}[htbp]
\caption{Results on MSCOCO source of POPE. The best performances are \textbf{bolded}. VCD and OPERA are two recently proposed training-free methods in CVPR24. * denotes our reproduced VCD results with Qwen-VL.}
\centering
\resizebox{\linewidth}{!}{%
\begin{tabular}{lllccc|c}
\hline
\textbf{Setting}                         & \textbf{Model}                & \textbf{Decoding} & Accuracy$\uparrow$ & Precision$\uparrow$ & Recall$\uparrow$ & F1 Score$\uparrow$  \\ \hline
\multirow{16}{*}{\textit{Random}}      & \multirow{4}{*}{LLaVA1.5}     & Vanilla           &$83.29_{(\pm0.35)}$ &$92.13_{(\pm0.54)}$ &$72.80_{(\pm0.57)}$ &$81.33_{(\pm0.41)}$   \\
                           &                               & VCD                &$87.73_{(\pm0.40)}$ &$91.42_{(\pm0.55)}$ &$83.28_{(\pm0.42)}$ &$87.16_{(\pm0.41)}$  \\
                           &                               & OPERA            &$89.17_{(\pm 0.15)}$ &$93.21_{(\pm 0.21)}$ &$85.20_{(\pm 0.37)}$ &$89.03_{(\pm 0.11)}$  \\
                           &                               & \cellcolor{gray!20} \textbf{Ours}             &\cellcolor{gray!20} $\textbf{91.10}_{(\pm 0.17)}$ &\cellcolor{gray!20} $93.69_{(\pm 0.25)}$ &\cellcolor{gray!20} $88.13_{(\pm 0.40)}$ &\cellcolor{gray!20} $\textbf{90.82}_{(\pm 0.16)}$  \\ \cline{3-7}
                          & \multirow{4}{*}{Qwen-VL}     & Vanilla           &$84.36_{(\pm0.48)}$ &$95.65_{(\pm0.43)}$ &$72.00_{(\pm0.32)}$ &$82.16_{(\pm0.51)}$  \\
                           &                               & VCD*               &$86.03_{(\pm0.13)}$ &$95.92_{(\pm0.35)}$ &$75.26_{(\pm0.13)}$ &$84.34_{(\pm0.10)}$  \\
                          &                               & OPERA             &$86.13_{(\pm 0.21)}$ &$97.54_{(\pm 0.37)}$ &$74.13_{(\pm 0.18)}$ &$84.24_{(\pm 0.22)}$  \\
                           &                               &\cellcolor{gray!20} \textbf{Ours}              &\cellcolor{gray!20} $\textbf{86.33}_{(\pm 0.25)}$ &\cellcolor{gray!20} $95.95_{(\pm 0.16)}$ &\cellcolor{gray!20} $75.86_{(\pm 0.22)}$ &\cellcolor{gray!20} $\textbf{84.74}_{(\pm 0.25)}$  \\ \cline{3-7}
                          & \multirow{4}{*}{InstructBLIP} & Vanilla           &$80.71_{(\pm0.73)}$ &$81.67_{(\pm0.67)}$ &$79.19_{(\pm1.14)}$ &$80.41_{(\pm0.80)}$  \\
                           &                               & VCD               &$84.53_{(\pm0.38)}$ &$88.55_{(\pm0.54)}$ &$79.32_{(\pm0.44)}$ &$83.68_{(\pm0.40)}$  \\
                           &                               & OPERA             &$89.86_{(\pm 0.24)}$ &$94.46_{(\pm 0.30)}$ &$85.33_{(\pm 0.47)}$ &$89.66_{(\pm 0.16)}$  \\
                            &                               &\cellcolor{gray!20} \textbf{Ours}              &\cellcolor{gray!20} $\textbf{90.30}_{(\pm 0.41)}$ &\cellcolor{gray!20} $92.54_{(\pm 0.28)}$ &\cellcolor{gray!20} $87.66_{(\pm 0.31)}$ &\cellcolor{gray!20} $\textbf{90.04}_{(\pm 0.19)}$  \\ \cline{3-7}
                           & \multirow{4}{*}{mPLUG-Owl2} & Vanilla           &$86.70_{(\pm 0.18)}$ &$91.73_{(\pm 0.45)}$ &$80.66_{(\pm 0.33)}$ &$85.84_{(\pm 0.56)}$  \\
                            &                               & VCD               &$88.13_{(\pm 0.24)}$ &$93.93_{(\pm 0.12)}$ &$81.53_{(\pm 0.45)}$ &$87.29_{(\pm 0.31)}$  \\
                            &                               & OPERA             &$86.90_{(\pm 0.26)}$ &$91.90_{(\pm 0.39)}$ &$80.93_{(\pm 0.17)}$ &$86.07_{(\pm 0.43)}$  \\
                           &                               &\cellcolor{gray!20}  \textbf{Ours}              &\cellcolor{gray!20} $\textbf{91.13}_{(\pm 0.26)}$ &\cellcolor{gray!20} $92.49_{(\pm 0.14)}$ &\cellcolor{gray!20} $89.53_{(\pm 0.36)}$ &\cellcolor{gray!20} $\textbf{90.98}_{(\pm 0.24)}$  \\ 
\cline{1-7}
\multirow{16}{*}{\textit{Popular}}      & \multirow{4}{*}{LLaVA1.5}     & Vanilla           &$81.88_{(\pm0.48)}$ &$88.93_{(\pm0.60)}$ &$72.80_{(\pm0.57)}$ &$80.06_{(\pm0.05)}$  \\                                    
                           &                               & VCD               &$85.38_{(\pm0.38)}$ &$86.92_{(\pm0.53)}$ &$83.28_{(\pm0.42)}$ &$85.06_{(\pm0.37)}$  \\
                           &                               & OPERA             &$86.00_{(\pm 0.33)}$ &$84.09_{(\pm 0.18)}$ &$88.80_{(\pm 0.44)}$ &$86.38_{(\pm 0.17)}$  \\
                           &                               & \cellcolor{gray!20} \textbf{Ours}              &\cellcolor{gray!20} $\textbf{87.06}_{(\pm 0.27)}$ &\cellcolor{gray!20} $84.84_{(\pm 0.13)}$ &\cellcolor{gray!20} $90.27_{(\pm 0.45)}$ &\cellcolor{gray!20} $\textbf{87.47}_{(\pm 0.39)}$  \\ \cline{3-7}
                          & \multirow{4}{*}{Qwen-VL}     & Vanilla           &$84.06_{(\pm0.18)}$ &$94.20_{(\pm0.43)}$ &$72.60_{(\pm0.45)}$ &$82.00_{(\pm0.23)}$  \\
                           &                               & VCD*               &$85.80_{(\pm0.07)}$ &$94.82_{(\pm0.10)}$ &$75.73_{(\pm0.19)}$ &$84.21_{(\pm0.09)}$  \\
                          &                               & OPERA             &$85.73_{(\pm 0.21)}$ &$96.52_{(\pm 0.15)}$ &$74.13_{(\pm 0.11)}$ &$83.86_{(\pm 0.14)}$  \\
                           &                               &\cellcolor{gray!20}  \textbf{Ours}              &\cellcolor{gray!20} $\textbf{85.96}_{(\pm 0.38)}$ &\cellcolor{gray!20} $94.40_{(\pm 0.11)}$ &\cellcolor{gray!20} $76.46_{(\pm 0.46)}$ &\cellcolor{gray!20} $\textbf{84.49}_{(\pm 0.21)}$  \\ \cline{3-7}
                          & \multirow{4}{*}{InstructBLIP} & Vanilla           &$78.22_{(\pm0.84)}$ &$77.87_{(\pm1.03)}$ &$78.85_{(\pm0.52)}$ &$78.36_{(\pm0.76)}$  \\
                           &                               & VCD               &$81.47_{(\pm0.42)}$ &$82.89_{(\pm0.64)}$ &$79.32_{(\pm0.44)}$ &$81.07_{(\pm0.39)}$  \\
                           &                               & OPERA             &$\textbf{83.43}_{(\pm 0.12)}$ &$81.21_{(\pm 0.46)}$ &$87.00_{(\pm 0.35)}$ &$\textbf{84.00}_{(\pm 0.24)}$  \\
                            &                               & \cellcolor{gray!20} \textbf{Ours}              &\cellcolor{gray!20} $79.93_{(\pm 0.23)}$ &\cellcolor{gray!20} $76.01_{(\pm 0.25)}$ &\cellcolor{gray!20} $87.46_{(\pm 0.37)}$ &\cellcolor{gray!20} $81.34_{(\pm 0.49)}$ \\  \cline{3-7}
                           & \multirow{4}{*}{mPLUG-Owl2} & Vanilla           &$83.66_{(\pm 0.37)}$ &$85.51_{(\pm 0.25)}$ &$81.06_{(\pm 0.48)}$ &$83.23_{(\pm 0.19)}$ \\
                            &                               & VCD               &$84.00_{(\pm 0.29)}$ &$80.57_{(\pm 0.13)}$ &$89.60_{(\pm 0.45)}$ &$84.85_{(\pm 0.21)}$ \\
                            &                               & OPERA              &$84.53_{(\pm 0.15)}$ &$87.59_{(\pm 0.38)}$ &$80.46_{(\pm 0.49)}$ &$83.87_{(\pm 0.22)}$  \\
                           &                               & \cellcolor{gray!20} \textbf{Ours}              &\cellcolor{gray!20} $\textbf{86.30}_{(\pm 0.03)}$ &\cellcolor{gray!20} $90.12_{(\pm 0.28)}$ &\cellcolor{gray!20} $81.53_{(\pm 0.14)}$ &\cellcolor{gray!20} $\textbf{85.61}_{(\pm 0.47)}$  \\ 
\cline{1-7}
\multirow{16}{*}{\textit{Adversarial}}      & \multirow{4}{*}{LLaVA1.5}     & Vanilla           &$78.96_{(\pm0.52)}$ &$83.06_{(\pm0.58)}$ &$72.75_{(\pm0.59)}$ &$77.57_{(\pm0.57)}$    \\
                           &                               & VCD               &$80.88_{(\pm0.33)}$ &$79.45_{(\pm0.29)}$ &$83.29_{(\pm0.43)}$ &$81.33_{(\pm0.34)}$  \\
                           &                               & OPERA             &$79.13_{(\pm 0.31)}$ &$74.41_{(\pm 0.23)}$ &$88.80_{(\pm 0.41)}$ &$80.97_{(\pm 0.14)}$  \\
                           &                               &\cellcolor{gray!20}  \textbf{Ours}              &\cellcolor{gray!20} $\textbf{81.50}_{(\pm 0.20)}$ &\cellcolor{gray!20} $78.20_{(\pm 0.44)}$ &\cellcolor{gray!20} $87.33_{(\pm 0.42)}$ &\cellcolor{gray!20} $\textbf{82.51}_{(\pm 0.33)}$  \\ \cline{3-7}
                          & \multirow{4}{*}{Qwen-VL}     & Vanilla           &$82.66_{(\pm0.30)}$ &$90.49_{(\pm0.33)}$ &$73.00_{(\pm0.50)}$ &$80.81_{(\pm0.37)}$  \\
                           &                               & VCD*               &$83.30_{(\pm0.39)}$ &$89.17_{(\pm0.45)}$ &$75.80_{(\pm0.39)}$ &$81.94_{(\pm0.39)}$  \\
                          &                               & OPERA             &$83.93_{(\pm 0.45)}$ &$92.55_{(\pm 0.22)}$ &$73.80_{(\pm 0.37)}$ &$82.12_{(\pm 0.18)}$  \\
                           &                               &\cellcolor{gray!20}  \textbf{Ours}              &\cellcolor{gray!20} $\textbf{84.23}_{(\pm 0.39)}$ &\cellcolor{gray!20} $92.89_{(\pm 0.12)}$ &\cellcolor{gray!20} $74.13_{(\pm 0.43)}$ &\cellcolor{gray!20} $\textbf{82.46}_{(\pm 0.28)}$  \\
                            \cline{3-7}
                          & \multirow{4}{*}{InstructBLIP} & Vanilla          &$75.84_{(\pm0.45)}$ &$74.30_{(\pm0.63)}$ &$79.03_{(\pm0.68)}$ &$76.59_{(\pm0.40)}$  \\
                           &                               & VCD               &$79.56_{(\pm0.41)}$ &$79.67_{(\pm0.59)}$ &$79.39_{(\pm0.50)}$ &$79.52_{(\pm0.38)}$  \\
                           &                               & OPERA             &$80.73_{(\pm 0.32)}$ &$77.31_{(\pm 0.46)}$ &$87.0_{(\pm 0.15)}$ &$81.87_{(\pm 0.23)}$  \\
                            &                               & \cellcolor{gray!20} \textbf{Ours}              &\cellcolor{gray!20} $\textbf{80.82}_{(\pm 0.24)}$ &\cellcolor{gray!20} $81.23_{(\pm 0.18)}$ &\cellcolor{gray!20} $82.91_{(\pm 0.38)}$ &\cellcolor{gray!20} $\textbf{82.06}_{(\pm 0.15)}$ \\ \cline{3-7}
                           & \multirow{4}{*}{mPLUG-Owl2} & Vanilla           &$81.73_{(\pm 0.44)}$ &$82.82_{(\pm 0.33)}$ &$80.06_{(\pm 0.18)}$ &$81.42_{(\pm 0.29)}$ \\
                            &                               & VCD               &$80.70_{(\pm 0.30)}$ &$75.94_{(\pm 0.17)}$ &$89.87_{(\pm 0.41)}$ &$82.33_{(\pm 0.24)}$ \\
                            &                               & OPERA              &$82.43_{(\pm 0.35)}$ &$83.48_{(\pm 0.28)}$ &$80.86_{(\pm 0.41)}$ &$82.15_{(\pm 0.14)}$  \\
                           &                               &\cellcolor{gray!20}  \textbf{Ours}              &\cellcolor{gray!20} $\textbf{84.40}_{(\pm 0.08)}$ &\cellcolor{gray!20} $86.49_{(\pm 0.37)}$ &\cellcolor{gray!20} $81.53_{(\pm 0.19)}$ &\cellcolor{gray!20} $\textbf{83.94}_{(\pm 0.61)}$  \\ 
\hline
\end{tabular}
}


\label{tab:pope}
\end{table*}

\begin{figure*}[htbp]
  \centering
  \begin{subfigure}[b]{0.24\textwidth}
    \includegraphics[width=\textwidth]{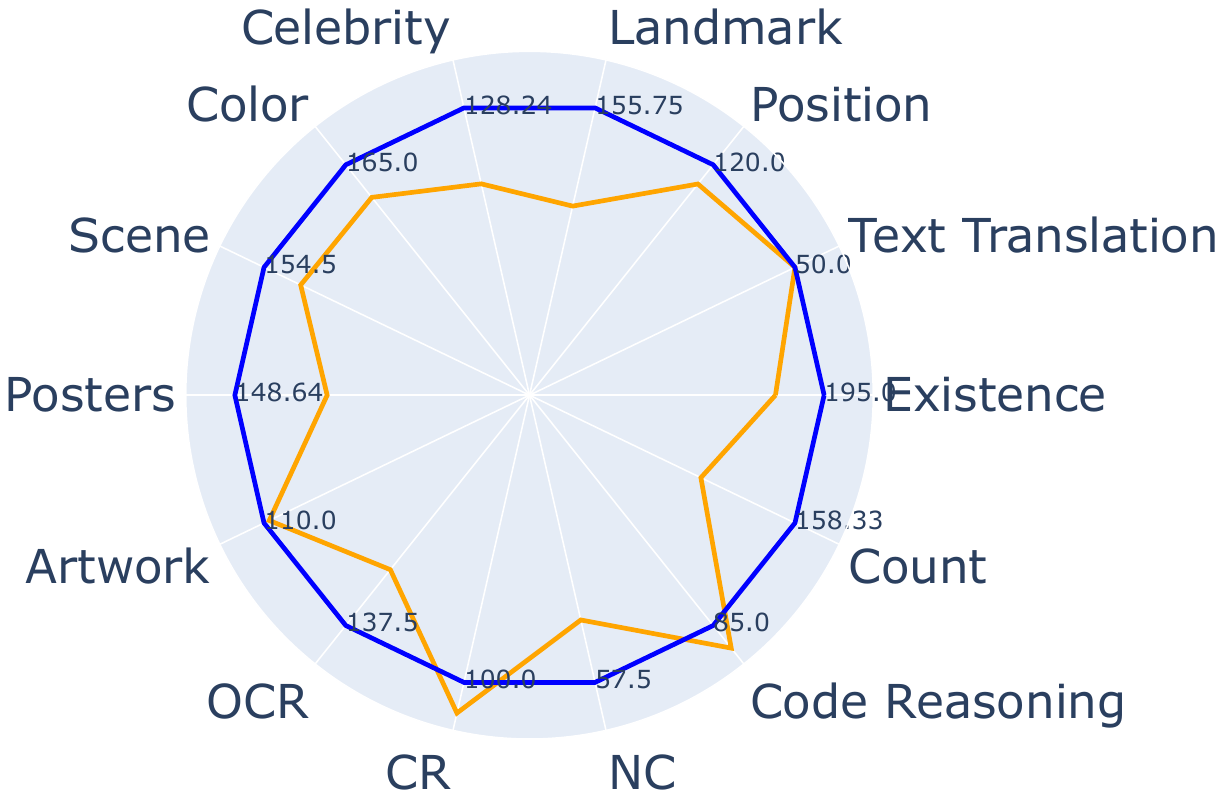}
  \end{subfigure}
  \begin{subfigure}[b]{0.24\textwidth}
    \includegraphics[width=\textwidth]{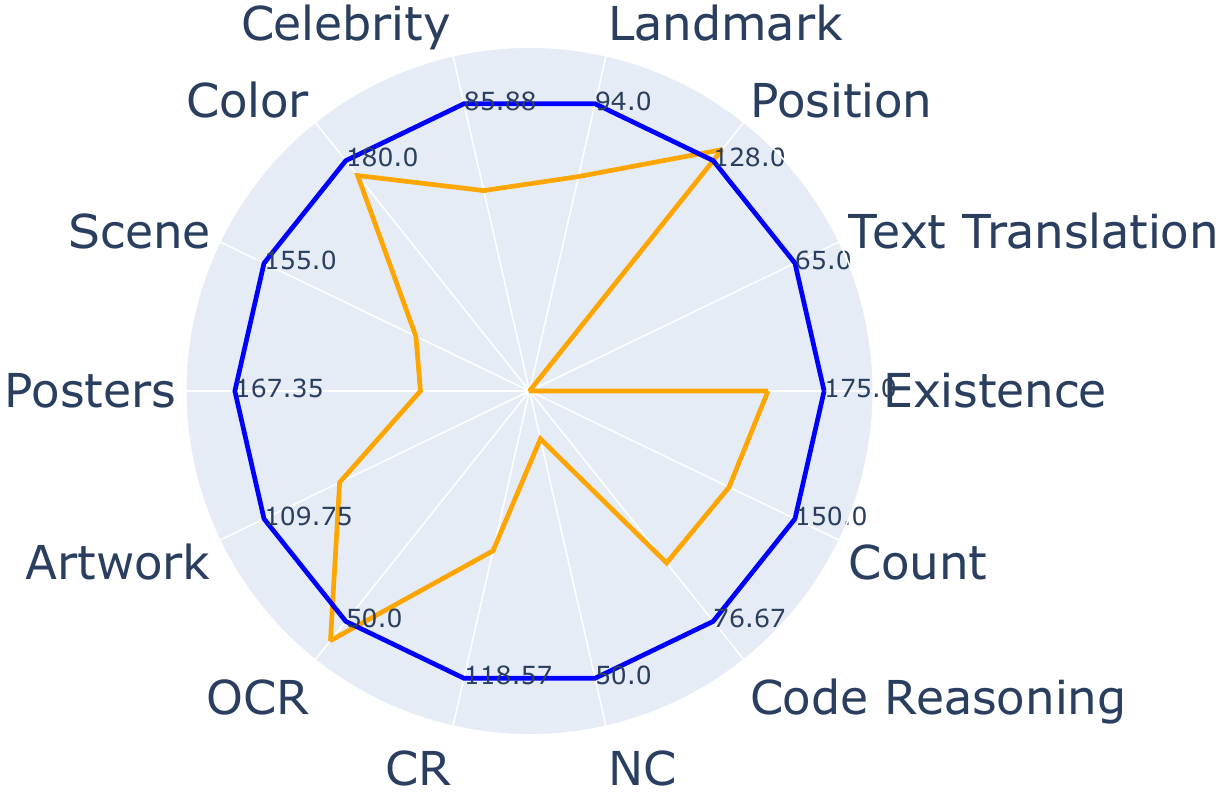}
  \end{subfigure}
  \begin{subfigure}[b]{0.24\textwidth}
    \includegraphics[width=\textwidth]{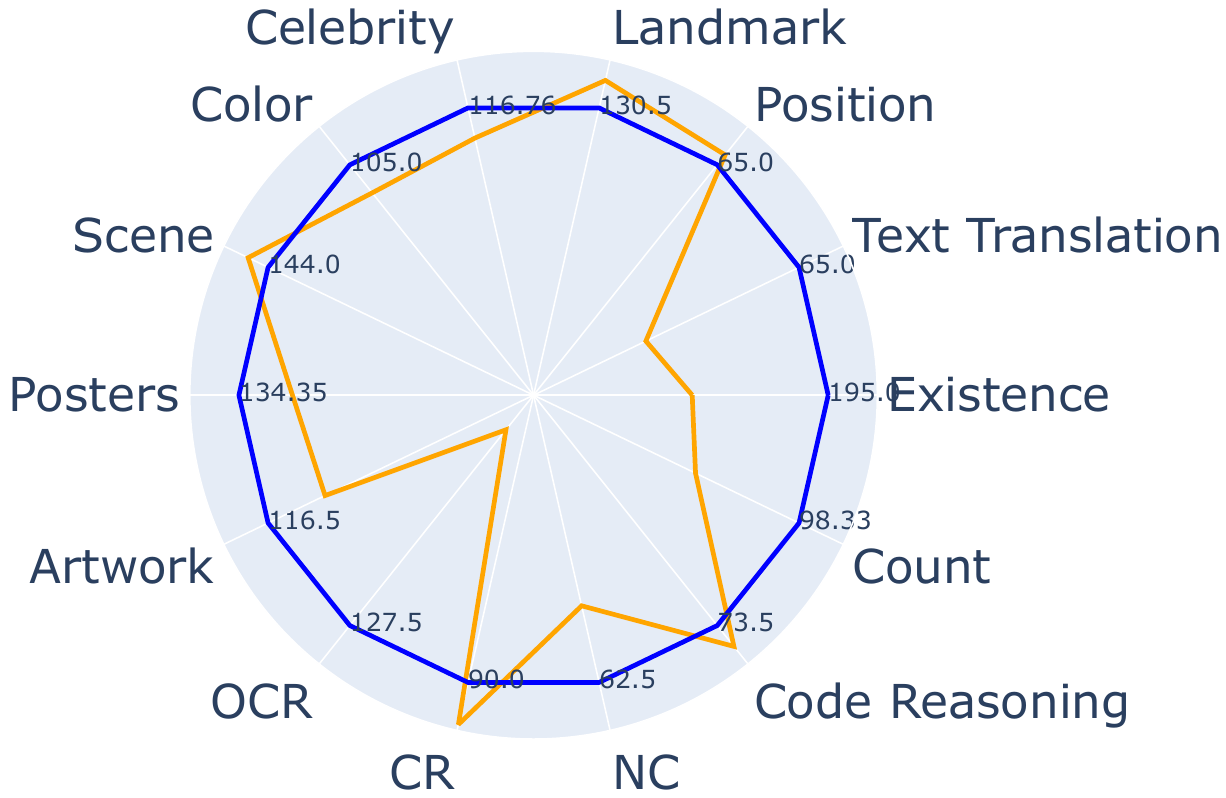}
  \end{subfigure}
  \begin{subfigure}[b]{0.24\textwidth}
    \includegraphics[width=\textwidth]{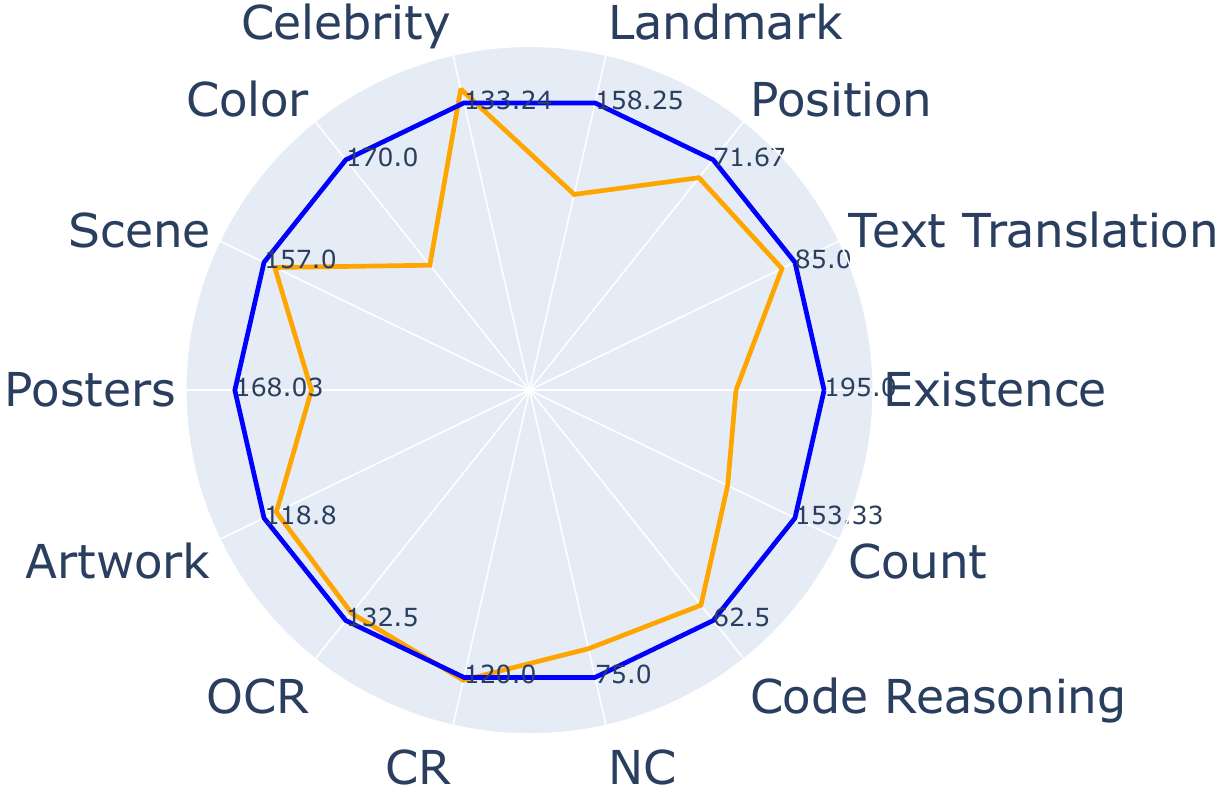}
  \end{subfigure}
  \caption{MME full set results on LLaVA-1.5, Qwen-VL, InstructBLIP, and mPLUG-Owl2 on 14 subtasks. 
  The orange lines represent the vanilla model and the blue lines denotes our MVP model.}
  \label{fig:mme_full}
\end{figure*}

\begin{table*}[h]
\caption{Results on the hallucination subset of MME. 
The best performances within each setting are \textbf{bolded}.}
\centering
\footnotesize
\resizebox{0.75\linewidth}{!}{%
\begin{tabular}{@{}llccccc@{}}
\toprule
\multirow{2}{*}{Model}        & \multirow{2}{*}{Decoding} & \multicolumn{2}{c}{\textbf{Object-level}}                                   & \multicolumn{2}{c}{\textbf{Attribute-level}}                               & \multicolumn{1}{c}{\multirow{2}{*}{Total Scores$\uparrow$}} \\
                              &                           & \multicolumn{1}{c}{\textit{Existence}$\uparrow$} & \multicolumn{1}{c}{\textit{Count}$\uparrow$} & \multicolumn{1}{c}{\textit{Position}$\uparrow$} & \multicolumn{1}{c}{\textit{Color}$\uparrow$} & \multicolumn{1}{c}{}                       \\ \midrule
\multirow{4}{*}{LLaVA1.5}     & Vanilla                   & $175.67$ & $124.67$ & $114.00$ & $151.00$ & $565.33$ \\
                              & VCD                       & $184.66$ & $138.33$ & $\textbf{128.67}$ & $153.00$ & $604.66$ \\ 

                              & OPERA                     & $185.00$ & $108.33$ & $111.67$ & $145.00$ & $550.00$ \\ 
                              & \textbf{Ours}                      & $\textbf{195.00}$ & $\textbf{158.33}$ & $120.00$ & $\textbf{165.00}$ & $\textbf{638.33}$ \\                              \midrule
\multirow{4}{*}{Qwen-VL}      & Vanilla                   & $155.00$ & $127.67$ & $131.67$ & $173.00$ & $587.33$ \\
                              & VCD                       & $156.00$ & $131.00$ & $128.00$ & $\textbf{181.67}$ & $596.67$ \\ 
                              & OPERA                     & $165.00$ & $145.00$ & $\textbf{133.33}$ & $180.00$ & $623.33$ \\                      
                              & \textbf{Ours}                      & $\textbf{175.00}$ & $\textbf{150.00}$ & $128.00$ & $180.00$ & $\textbf{633.00}$ \\ 
                              \midrule
\multirow{4}{*}{InstructBLIP} & Vanilla                   & $141.00$ & $75.33$ & $\textbf{66.67}$ & $97.33$ & $380.33$ \\
                              & VCD                       & $168.33$ & $92.33$ & $64.00$ & $\textbf{123.00}$ & $447.67$ \\ 
                              & OPERA                     & $156.00$ & $78.33$ & $55.00$ & $95.00$ & $384.33$ \\ 
                              & \textbf{Ours}                      & $\textbf{195.00}$ & $\textbf{98.33}$ & $65.00$ & $105.0$ & $\textbf{456.67}$ \\ 
                              \midrule
                              
\multirow{4}{*}{mPLUG-Owl2}   & Vanilla                   & $160.00$ & $130.00$ & $68.33$ & $123.33$ & $481.66$ \\
                              & VCD                       & $170.00$ & $\textbf{155.00}$ & $71.67$ & $141.67$ & $538.34$ \\ 
                              & OPERA                     & $173.33$ & $150.00$ & $\textbf{85.00}$ & $138.33$ & $546.66$ \\ 
                              & \textbf{Ours}                      & $\textbf{195.00}$ & $153.33$ & $71.67$ & $\textbf{170.00}$ & $\textbf{589.99}$ \\ 
                              
\bottomrule
\end{tabular}
}
\label{tab:mme}
\end{table*}

\section{Experiment}


\subsection{Evaluation Benchmarks}
Following previous works \cite{leng2023mitigating,huang2023opera}, we use the following two benchmarks POPE and MME.

\noindent\textbf{POPE} the Polling-based Object Probing Evaluation \cite{li2023evaluating}. In this benchmark, LVLMs are queried to determine whether a specific object is present in the provided image. 
It encompasses three distinct settings: random, popular, and adversarial, each differing in the construction of negative samples. 
The POPE benchmark aggregates data from three distinct sources: MSCOCO \cite{lin2014microsoft}, A-OKVQA \cite{schwenk2022okvqa}, and GQA \cite{hudson2019gqa}. It involves 500 images from each dataset under each sampling setting.
The performance is gauged using four key metrics: Accuracy, Precision, Recall, and F1.

\noindent \textbf{MME} \cite{fu2024mme} acts as a comprehensive benchmark designed to evaluate LVLMs across a range of dimensions. It is composed of ten perception-related subtasks and four cognition-focused ones. 
In the experiments, we evaluate the full dataset. In addition, we take into account the existence and count subsets for the inspection of object-level hallucination, along with the position and color subsets for attribute-level hallucination evaluation. 
The combined metric of accuracy and accuracy+ is used to quantify the performance as per the official implementation.


\subsection{Evaluation LVLM and Baselines}

\textbf{LVLMs}. To comprehensively evaluate our model and have a fair comparison with previous works, we experiment with our proposed MVP on four state-of-the-art LVLMs, including LLaVA1.5, Qwen-VL, InstructBLIP, and mPLUG-Owl2. 
{All four LVLMs are based on 7B LLM backbone models.

\noindent \textbf{Baselines}. 
To verify the effectiveness of our framework, we compare MVP with the vanilla LVLMs and two recent training-free methods, including VCD and OPERA. 
In our main experiments, for fair comparison, vanilla, VCD, and our MVP all adopt the decoding strategy of direct sampling.
In addition, OPERA introduces a penalty term on the model logits during the beam-search decoding to mitigate the over-trust issue.

\begin{table}[t]
\caption{Performance comparison of different views. Here ``Bottom" means bottom-up perspective, and ``Top" indicates top-down view. The experiments are conducted on ``Adversarial" MSCOCO part of POPE using LLaVA1.5 model.}
\centering
\scalebox{0.65}{
\begin{tabular}{ccc|cccc}
\hline
\textbf{Regular} & \textbf{Bottom} & \textbf{Top} & \textbf{Accuracy}  & \textbf{Precision}& \textbf{Recall}   & \textbf{F1} \\
\hline
- & - & - & 79.33 & 74.88      & 88.26  & 81.02   \\
\checkmark & ~ & ~ & 80.36 & 78.09      & 84.40  & 81.13   \\
~ & \checkmark & ~ &  80.60 & 78.15      & 84.93  & 81.40   \\
~ & ~ & \checkmark &  80.73 & 78.35      & 84.93  & 81.51   \\
\checkmark & \checkmark & ~ &  80.60 & 78.06      & 85.13  & 81.44   \\
\checkmark & ~ & \checkmark & 80.86 & 78.26     & 85.47  & 81.71   \\
~ & \checkmark & \checkmark & 80.90 & 77.80     & 86.46  & 81.91  \\
\checkmark & \checkmark & \checkmark     & \textbf{81.50} & 78.20      & 87.33  & \textbf{82.51}  \\ 
\hline
\end{tabular}
}
\label{tab:dataset_performance_comparison}
\end{table}

\subsection{Experiment Results}

\textbf{Results on POPE.}
Table \ref{tab:pope} summarizes the experimental results on the MSCOCO part of POPE benchmark, including experiments under random, popular, and adversarial settings. The results of A-OKVQA and GQA are presented in the Appendix. 
Specifically, under different settings, our method significantly surpasses the vanilla model's performance across all LVLMs. For example, with LLaVA1.5, MVP achieves an average improvement of 15.9 in Accuracy and 21.84 in F1 score across random, popular, and adversarial settings. 
For LLaVA1.5, Qwen-VL, and InstructBLIP, the improvement in F1 scores is mainly due to an increase in recall,
while in mPLUG-Owl2, the increase comes from the simultaneous improvement of precision and recall. Furthermore, compared to VCD and OPERA, our method still achieves better results in most cases. These results demonstrate the effectiveness of our MVP in alleviating hallucinations.


\noindent\textbf{Results on MME Hallucination Subset.} 
We further evaluate our method on a subset of MME, which includes object-level hallucinations and attribute-level hallucinations. The results in Table \ref{tab:mme} demonstrate that our method significantly improves the performance of all LVLMs in addressing object-level and attribute-level hallucinations.
Additionally, we compare our method with the recent strong methods VCD and OPERA, and our approach still exhibits better overall performance. 

\noindent \textbf{Results on MME Full Set}. 
As depicted in Figure 6, we test our method on the complete MME set to assess the overall capability of models. 
Four models with our MVP (blue lines) present significant improvement compared to the vanilla models on most evaluation subsets. 
This can be attributed to our method's introduction of multi-view information and multi-path reasoning, 
allowing LVLMs to comprehensively understand visual elements in the images and carefully consider the certainty of potential answers, thereby enhancing the LVLMs' capabilities in downstream tasks.

\subsection{Ablation Study}


\subsubsection{Effectiveness of Multi-view Caption}
Table \ref{tab:dataset_performance_comparison} presents the performance from different perspectives. 
The first row presents the performance without using any additional caption information, while rows 2-4 respectively use a single perspective. 
The improvement is more pronounced with more perspectives involved.
These results have confirmed that multi-view information can contribute to more comprehensive image understanding, thus mitigating the hallucinations in LVLMs.

\begin{table}[t]
\caption{The transferability of captions. With captions from LLaVA1.6, our models can achieve further improvement.}
\centering
\resizebox{0.90\linewidth}{!}{
\begin{tabular}{lc|ccc|c}
\toprule
\textbf{Model}        &  \textbf{Caption}         &  \textbf{Accuracy} & \textbf{Precision} & \textbf{Recall}   & \textbf{F1}     \\ \midrule
\multirow{3}{*}{LLaVA1.5}    &  \ding{55}        &  91.10 &  93.69 &  88.13 &  90.82 \\
              &   \checkmark         & 92.36 & 94.04 & 90.46 & 92.22 \\
              &         & \textcolor[RGB]{0,128,0}{(+1.26)} & \textcolor[RGB]{0,128,0}{(+0.35)} & \textcolor[RGB]{0,128,0}{(+2.33)} & \textcolor[RGB]{0,128,0}{(+1.40)} \\ \cline{2-6}
\multirow{3}{*}{Qwen-VL}   &  \ding{55}          &  86.33 & 95.95 &75.86 & 84.74 \\
              &  \checkmark          &  86.56 & 95.97 & 76.33 & 85.03 \\
              &        & \textcolor[RGB]{0,128,0}{(+0.23)} & \textcolor[RGB]{0,128,0}{(+0.02)} & \textcolor[RGB]{0,128,0}{(+0.47)} & \textcolor[RGB]{0,128,0}{(+0.29)} \\  \cline{2-6}
\multirow{3}{*}{InstructBLIP} &  \ding{55}         & 90.30 & 92.54 & 87.66 &  90.04 \\
              &  \checkmark     &  91.33 & 92.80 & 89.73 &  91.24 \\
              &       & \textcolor[RGB]{0,128,0}{(+1.03)} & \textcolor[RGB]{0,128,0}{(+0.26)} & \textcolor[RGB]{0,128,0}{(+2.07)} & \textcolor[RGB]{0,128,0}{(+1.20)} \\ \cline{2-6}
\multirow{3}{*}{mPLUG-Owl2}    &  \ding{55}       & 91.13 & 92.49 & 89.53 & 90.98 \\
              &  \checkmark        & 91.26 &92.98 & 89.85 & 91.39\\
              &       & \textcolor[RGB]{0,128,0}{(+0.13)} & \textcolor[RGB]{0,128,0}{(+0.49)} & \textcolor[RGB]{0,128,0}{(+0.32)} & \textcolor[RGB]{0,128,0}{(+0.41)} \\ \bottomrule
\end{tabular}
}

\label{tab:transferability}
\end{table}

\subsubsection{The transferability of captions.}
Intuitively, the quality of the captions has a direct impact on the model performance. Therefore, in this study, we explore the transferability of captions.
Specifically, we employ a more powerful open-source model LLaVA1.6 \cite{liu2024llavanext} to generate three-perspective captions for images in the random part of POPE MSCOCO, and use these captions for our model. 
As depicted in Table \ref{tab:transferability}, with better captions, our method provides further and stable improvements across four LVLMs. This result also confirms the significance of the multi-view information for alleviating hallucinations in LVLMs, as well as the plug-and-play flexibility of our method.

\subsubsection{Multi-path Reasoning}

Considering that we have multiple views of captions, we only adopt the regular perspective in this ablation study. 
We experiment on the random and adversarial part of the POPE MSCOCO benchmark, as shown in Figure \ref{fig:multiple_paths}.  

We first conduct experiments on Top-$K$ in Equation 5. 
As K increases from 1 to 5, 
peak performance is observed at K equals 3. When K becomes larger, the performance does not improve. This is due to the decoding path extending from the first tokens with minuscule probabilities do not provide any beneficial information.


Secondly, we explore a new aggregation strategy MVP-Max. Instead of accumulating the certainty scores in Equation 5, the potential answer with maximum certainty score among all paths is chosen as the final answer.
It can be seen that after using MVP-Max, the final performance of the model decreases significantly. 
This demonstrates the effectiveness of our aggregation strategy. 

Finally, we explore removing $M_j$ in Equation 5, we found that relying solely on the certainty of tokens can damage the stability and effectiveness of our model.



\begin{figure}
\setlength{\belowcaptionskip}{-1.5em}
\centering
\includegraphics[width=0.48\textwidth]{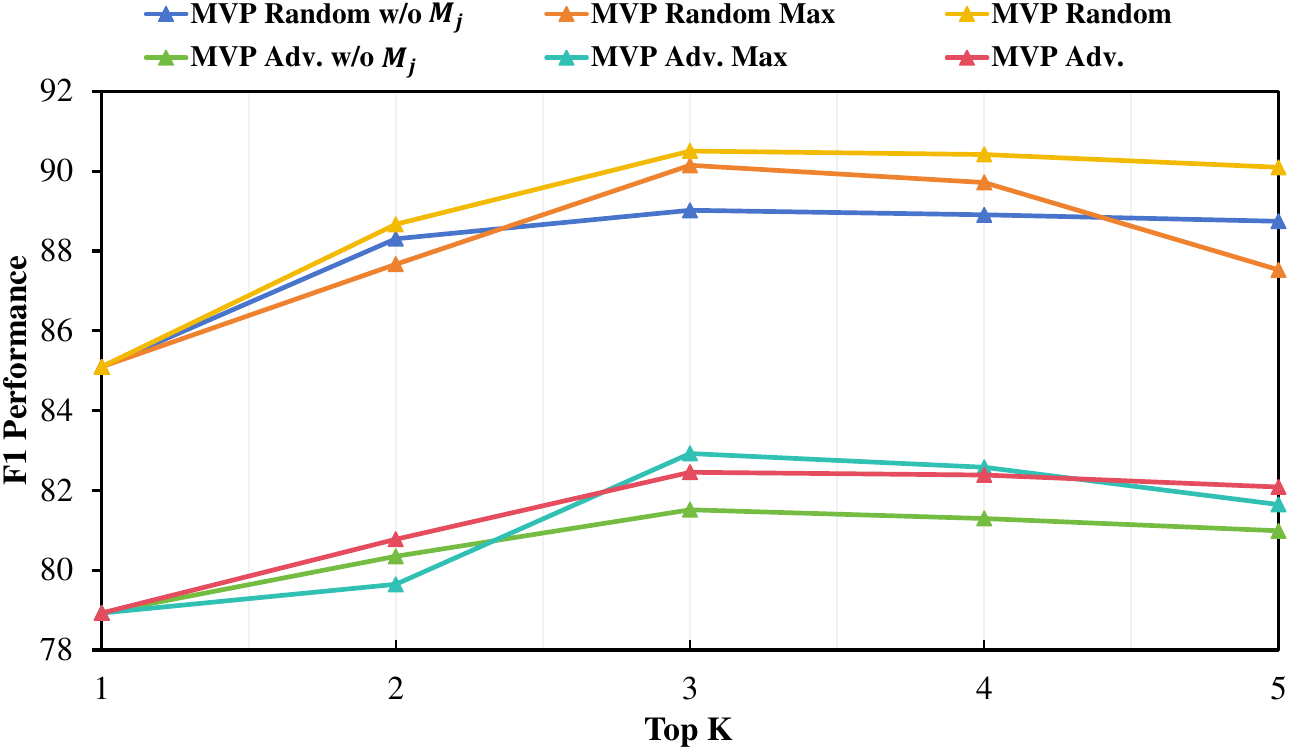}
\caption{The ablation study of multi-path variants.}  
\label{fig:multiple_paths}
\end{figure}

\subsection{Decoding Strategy}

In this section, we analyze the impact of different decoding strategies on our method. Specifically, we investigate five decoding methods. 
Notably, nucleus sampling is used in our main experiments for a fair comparison with recent methods.
Our MVP can further enhance the performance with training-free decoding methods such as VCD and OPERA, as shown in Table \ref{tab:decoding_strategy}. 
We can observe that using beam search as the decoding strategy performs the best accuracy on the random setting, while OPERA achieves the most significant accuracy on the adversarial part. 
These results also imply that our method is a novel plug-and-play approach, which can be flexibly integrated with other techniques.

\begin{table}[tp]
\caption{Ablation of different decoding strategies using LLaVA1.5. Experiments are performed on the MSCOCO source of POPE metric. The best performances are \textbf{bolded}.}
\centering
\resizebox{0.96\linewidth}{!}{%
\begin{tabular}{llccc|c}
\hline
\textbf{Setting}                         &  \textbf{Decoding} & Accuracy$\uparrow$ & Precision$\uparrow$ & Recall$\uparrow$ & F1 Score$\uparrow$  \\ \hline
\multirow{5}{*}{\textit{Random}}        & Greedy           &$91.16$ &$93.63$ &$88.31$ &$90.89$   \\
                                         & Nucleus              &$91.10$ &$93.69$ &$88.13$ &$90.82$  \\
                                         & Beam Search         &$\textbf{92.36}$ &$94.04$ &$90.46$ &$\textbf{92.21}$    \\
                                        & VCD              &$89.73$ &$95.77$ &$83.13$ &$89.01$  \\
                                        & OPERA         &$89.90$ &$96.15$ &$83.17$ &$89.19$  \\
\cline{2-6}
\cline{2-6}
\multirow{5}{*}{\textit{Adversarial}}        & Greedy           &$81.63$ &$78.34$ &$87.39$ &$82.62$   \\
                                         & Nucleus              &$81.50$ &$78.20$ &$87.33$ &$82.51$  \\
                                         & Beam Search           &$81.86$ &$77.63$ &$89.53$ &$\textbf{83.16}$  \\
                                        & VCD              &$82.56$ &$82.20$ &$83.13$ &$82.66$  \\
                                        & OPERA         &$\textbf{82.65}$ &$82.31$ &$83.22$ &$82.86$  \\
\hline
\end{tabular}
}
\label{tab:decoding_strategy}
\end{table}

\section{Conclusion}

In this paper, we propose a novel training-free framework MVP to reduce hallucinations by making the most of the innate capabilities of the LVLMs through Multi-View Multi-Path Reasoning. Specifically, we
devise a multi-view information-seeking strategy to perceive the
intricate details of the image information, which contributes to comprehensive image understanding. Furthermore, we propose multi-path reasoning to quantify and aggregate the certainty scores for each potential answer and finally decide the output answer. With the multi-view multi-path reasoning, our method effectively alleviates hallucinations in LVLMs.



\bibliography{custom}

\appendix



\section{Implementation Details}
In our paper, we adopt three different views to capture the information from the image.
For ``Bottom-up'' perspective, we use following prompt: \textit{``Through a systematic examination of the image at the pixel level and by analyzing various visual features, such as shape, color, and texture, along with employing object detection techniques, describe this
image in details.''} In addition, we use \textit{``Describe this image in details''} for the regular caption. 
The prompt for top-down perspective has been described in Section 3.3.  
In addition, to generate these multi-view captions, a temperature of 0.9 and a top-p parameter of 0.95 are set to guarantee diversity.
{$K$ in Equation 5 is set to 3. }
$\alpha$ in Equation 6 is set to 0.4, 0.2, and 0.4 for ``bottom-up'', ``regular'', and ``top-down'' perspectives, considering that the ``bottom-up'' and ``top-down'' perspectives bring more beneficial information. We do not tune $\alpha$ much as the aggregation from multiple decoding paths inherently presents certain robustness.

\section{Related Work}

\subsection{Hallucination in LVLMs}

Recently, the potential hallucination, conflict, and safety issues \cite{qu2024alleviating,qu2024mitigating,liu2024survey,lu2024twin,su2024conflictbank} in large models have garnered considerable attention, mainly due to the direct impact on the downstream applications. 
In LVLMs, the term ``hallucination'' refers to models that generate seemingly plausible outputs inclusive of objects, attributes, or relations that do not correspond with the images.

Regarding hallucination mitigation, the primary focus of most current methods has been to enhance the quality of the supervised fine-tuning or reinforcement learning data. 
VIGC \cite{wang2024vigc} presents a component to correct visual instructions with the aim to minimize hallucinations generated from lengthy sequences. 
LRV \cite{liu2023mitigating} attempt to alleviate hallucinations by developing 
expansive and diverse SFT data. For methods based on reinforcement learning, 
LLaVA-RLHF \cite{sun2023aligning} is the pioneer in applying Reinforcement Learning with Human Feedback (RLHF) to mitigate hallucination in LVLMs. 
RLHF-V \cite{yu2023rlhf} further develops a fine-grained correctional human feedback. Considering the instability and training difficulty of RLHF, \citet{zhao2023beyond} employ Direct Preference Optimization (DPO) and build a hallucination-aware dataset for alleviating hallucination. 
Although these methods have achieved significant improvements, they inevitably introduce a large training cost and are prone to overfitting to the training data. 
Instead, there are also training-free works aiming to solve the hallucination without introducing training cost. VCD \cite{leng2023mitigating} contrasts the output distributions derived from original and distorted visual inputs, aiming to recalibrate the model's excessive dependence on unimodal priors and statistical biases. OPERA \cite{huang2023opera} introduces a penalty term to the model logits during the beam-search decoding to alleviate the overconfidence problem, complemented by a rollback strategy.
In this paper, we focus on the training-free paradigm for mitigating
hallucination in LVLMs.

\subsection{Training-free Decoding Strategy}

As the recent training-free methods for mitigating hallucination focus on the decoding process, we describe several decoding strategies here. 
Decoding strategies in language models are instrumental in guiding how these models produce text. They are a significant factor in influencing the quality, relevance, and coherence of the output. Greedy decoding takes the simplest approach, choosing the most probable next word at every step. Despite its speed and computational efficiency, this method often results in repetitive and monotonous text. Conversely, beam search offers a more advanced technique that maintains a predetermined number of hypotheses at each step, elaborating on them to identify a more optimum sequence. In nucleus sampling, a flexible range of words is considered, which accumulate to achieve the given probability p. More recently, there are two methods specifically proposed for mitigating hallucinations. VCD adopts contrastive decoding to calibrate the model’s output distribution. 
OPERA introduces a penalty term on
the model logits during the beam-search decoding to mitigate the over-trust issue. 
In this paper, we focus on the certainty of the answer token during the decoding process, which does not conflict with designing different decoding paths. Thus, our MVP framework is plug-and-play and can further combine with the above decoding strategies.

\section{Qualitative Results}

To qualitatively verify the effectiveness of our method on downstream tasks, we presented two examples from the MSCOCO POPE and MME datasets. 
As illustrated in Figures \ref{fig:example2}, in both examples, MVP is able to accurately address questions. In the top figure where the chair is in the top right corner and not fully visible, our MVP comprehensively captures multi-view information, thus contributing to identifying this chair. 
In the bottom figure where distant people appear blurry, our MVP effectively decides the accurate number of people with multi-path reasoning.
Through these two direct comparisons, our method answers questions more precisely than the currently existing strong baselines, significantly reducing hallucinations in LVLM models.

\begin{figure}
\centering
\includegraphics[width=7.5cm]{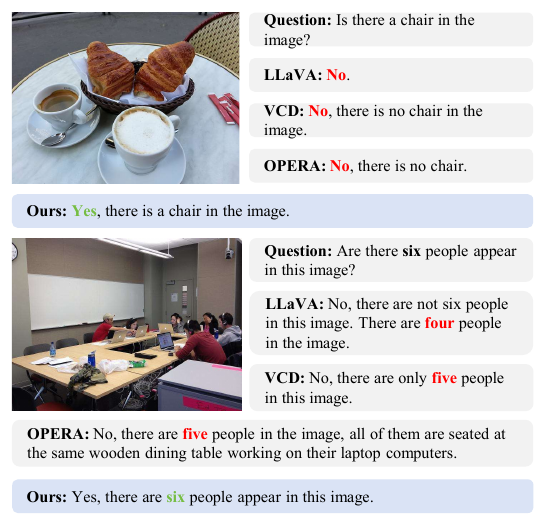}
\caption{Two representative examples from POPE and MME datasets. The qualitative results of LLaVA 1.5, VCD, OPERA, and our proposed MVP.}  
\label{fig:example2}
\end{figure}

\section{More Model Performance}
To further demonstrate the effectiveness of our proposed MVP, we conduct experiments on POPE based on AOKVQA and GQA with random, popular, and adversarial settings, respectively. The experiment settings are the same as the main experiment in the MSCOCO.
The results are shown in Tables \ref{tab:pope_aokvqa} and \ref{tab:pope_gqa}. It is obvious that our proposed MVP has greatly improved from these two tables compared with the baseline models.  
Specifically, under different settings, our method significantly surpasses the vanilla model's performance across all LVLMs. For example, with LLaVA1.5, MVP achieves an average improvement of 4.38 in Accuracy and 6.29 in F1 score across random, popular, and adversarial settings on AOKVQA. 
For LLaVA1.5, InstructBLIP and mPLUG-Owl2, the improvement in F1 and Accuracy scores is mainly due to an increase in precision,
while in Qwen-VL, the increase comes from the simultaneous improvement of precision and recall. These results demonstrate the effectiveness of our MVP in alleviating hallucinations.

\begin{table*}[!ht]
\centering
\resizebox{0.80\linewidth}{!}{%
\begin{tabular}{lllccc|c}
\hline
\textbf{Setting}                         & \textbf{Model}                & \textbf{Decoding} & Accuracy$\uparrow$ & Precision$\uparrow$ & Recall$\uparrow$ & F1 Score$\uparrow$  \\ \hline
\multirow{8}{*}{\textit{Random}}      & \multirow{2}{*}{LLaVA1.5}     & Vanilla           &$83.45_{(\pm0.48)}$ &$87.24_{(\pm0.68)}$ &$78.36_{(\pm0.54)}$ &$82.56_{(\pm0.50)}$   \\
                           &                               & \textbf{Ours}              & $\textbf{91.20}_{(\pm0.48)}$ & $89.06_{(\pm0.68)}$ & $93.93_{(\pm0.54)}$ & $\textbf{91.43}_{(\pm0.50)}$  \\ \cline{3-7}
                          & \multirow{2}{*}{Qwen-VL}     & Vanilla          &$86.67_{(\pm0.48)}$ &$93.16_{(\pm0.55)}$ &$79.16_{(\pm0.59)}$ &$85.59_{(\pm0.53)}$  \\
                           &                               & \textbf{Ours}              &$\textbf{88.16}_{(\pm 0.45)}$ &$94.34_{(\pm 0.26)}$ &$81.20_{(\pm 0.19)}$ &$\textbf{87.28}_{(\pm 0.06)}$  \\ \cline{3-7}
                          & \multirow{2}{*}{InstructBLIP} & Vanilla           &$80.91_{(\pm0.34)}$ &$77.97_{(\pm0.59)}$ &$86.16_{(\pm0.88)}$ &$81.86_{(\pm0.32)}$  \\
                            &                               & \textbf{Ours}              &$\textbf{88.60}_{(\pm 0.16)}$ &$90.95_{(\pm 0.32)}$ &$85.73_{(\pm 0.50)}$ &$\textbf{88.26}_{(\pm 0.07)}$  \\ \cline{3-7}
                           & \multirow{2}{*}{mPLUG-Owl2} & Vanilla           &$77.63_{(\pm 0.14)}$ &$70.46_{(\pm 0.29)}$ &$95.13_{(\pm 0.42)}$ &$80.96_{(\pm 0.11)}$  \\
                           &                               & \textbf{Ours}              &$\textbf{90.30}_{(\pm 0.33)}$ &$88.92_{(\pm 0.44)}$ &$92.07_{(\pm 0.11)}$ &$\textbf{90.47}_{(\pm 0.28)}$  \\ 
\cline{1-7}
\multirow{8}{*}{\textit{Popular}}      & \multirow{2}{*}{LLaVA1.5}     & Vanilla           &$79.90_{(\pm0.33)}$ &$80.85_{(\pm0.31)}$ &$78.36_{(\pm0.54)}$ &$79.59_{(\pm0.37)}$   \\                      
                           &                               & \textbf{Ours}              &$\textbf{84.60}_{(\pm 0.23)}$ &$79.19_{(\pm 0.37)}$ &$93.87_{(\pm 0.08)}$ &$\textbf{85.91}_{(\pm 0.49)}$  \\ \cline{3-7}
                          & \multirow{2}{*}{Qwen-VL}     & Vanilla           &$85.56_{(\pm0.35)}$ &$90.44_{(\pm0.56)}$ &$79.53_{(\pm0.84)}$ &$84.63_{(\pm0.42)}$  \\
                           &                               & \textbf{Ours}              &$\textbf{87.30}_{(\pm 0.15)}$ &$92.48_{(\pm 0.33)}$ &$81.20_{(\pm 0.27)}$ &$\textbf{86.47}_{(\pm 0.40)}$  \\ \cline{3-7}
                          & \multirow{2}{*}{InstructBLIP} & Vanilla          &$76.19_{(\pm0.80)}$ &$72.16_{(\pm0.69)}$ &$85.28_{(\pm0.79)}$ &$78.17_{(\pm0.73)}$  \\
                            &                               & \textbf{Ours}              &$\textbf{78.16}_{(\pm0.80)}$ &$74.90_{(\pm0.69)}$ &$84.73_{(\pm0.79)}$ &$\textbf{79.51}_{(\pm0.73)}$ \\  \cline{3-7}
                           & \multirow{2}{*}{mPLUG-Owl2} & Vanilla           &$72.06_{(\pm 0.15)}$ &$65.16_{(\pm 0.30)}$ &$94.80_{(\pm 0.05)}$ &$77.24_{(\pm 0.33)}$ \\
                           &                               & \textbf{Ours}              &$\textbf{83.30}_{(\pm 0.28)}$ &$78.56_{(\pm 0.47)}$ &$91.60_{(\pm 0.13)}$ &$\textbf{84.58}_{(\pm 0.22)}$  \\ 
\cline{1-7}
\multirow{8}{*}{\textit{Adversarial}}      & \multirow{2}{*}{LLaVA1.5}  & Vanilla       &$74.04_{(\pm0.34)}$ &$72.08_{(\pm0.53)}$ &$78.49_{(\pm0.38)}$ &$75.15_{(\pm0.23)}$    \\
                           &                               & \textbf{Ours}              &$\textbf{74.70}_{(\pm 0.12)}$ &$67.75_{(\pm 0.45)}$ &$94.27_{(\pm 0.36)}$ &$\textbf{78.84}_{(\pm 0.09)}$  \\ \cline{3-7}
                          & \multirow{2}{*}{Qwen-VL}     & Vanilla          &$79.57_{(\pm0.31)}$ &$79.77_{(\pm0.34)}$ &$79.23_{(\pm0.73)}$ &$79.50_{(\pm0.38)}$  \\
                           &                               & \textbf{Ours}              &$\textbf{81.60}_{(\pm 0.22)}$ &$81.85_{(\pm 0.38)}$ &$81.20_{(\pm 0.07)}$ &$\textbf{81.53}_{(\pm 0.54)}$  \\
                            \cline{3-7}
                          & \multirow{2}{*}{InstructBLIP} & Vanilla          &$70.71_{(\pm0.76)}$ &$65.91_{(\pm0.74)}$ &$85.83_{(\pm0.80)}$ &$75.56_{(\pm0.57)}$ \\
                            &                               & \textbf{Ours}              &$\textbf{75.03}_{(\pm 0.16)}$ &$73.75_{(\pm 0.31)}$ &$85.25_{(\pm 0.08)}$ &$\textbf{79.08}_{(\pm 0.29)}$ \\ \cline{3-7}
                           & \multirow{2}{*}{mPLUG-Owl2} & Vanilla           &$55.13_{(\pm 0.14)}$ &$53.26_{(\pm 0.42)}$ &$83.73_{(\pm 0.24)}$ &$65.11_{(\pm 0.33)}$ \\
                           &                               & \textbf{Ours}              &$\textbf{73.43}_{(\pm 0.19)}$ &$67.32_{(\pm 0.48)}$ &$91.07_{(\pm 0.11)}$ &$\textbf{77.42}_{(\pm 0.37)}$  \\ 
\hline
\end{tabular}
}
\caption{Results on AOKVQA source of POPE benchmark. The best performances are \textbf{bolded}.}
\label{tab:pope_aokvqa}
\end{table*}

\begin{table*}
\centering
\resizebox{0.80\linewidth}{!}{%
\begin{tabular}{lllccc|c}
\hline
\textbf{Setting}                         & \textbf{Model}                & \textbf{Decoding} & Accuracy$\uparrow$ & Precision$\uparrow$ & Recall$\uparrow$ & F1 Score$\uparrow$  \\ \hline
\multirow{8}{*}{\textit{Random}}      & \multirow{2}{*}{LLaVA1.5}     & Vanilla           &$83.73_{(\pm0.27)}$ &$87.16_{(\pm0.39)}$ &$79.12_{(\pm0.35)}$ &$82.95_{(\pm0.28)}$   \\
                           &                               & \textbf{Ours}              &$\textbf{90.20}_{(\pm 0.17)}$ &$87.64_{(\pm 0.29)}$ &$93.60_{(\pm 0.38)}$ &$\textbf{90.52}_{(\pm 0.07)}$  \\ \cline{3-7}
                          & \multirow{2}{*}{Qwen-VL}     & Vanilla           &$80.97_{(\pm0.32)}$ &$88.07_{(\pm0.34)}$ &$71.64_{(\pm0.57)}$ &$79.01_{(\pm0.40)}$ \\
                           &                               & \textbf{Ours}              &$\textbf{83.57}_{(\pm 0.24)}$ &$91.10_{(\pm 0.35)}$ &$74.40_{(\pm 0.10)}$ &$\textbf{81.91}_{(\pm 0.44)}$  \\ \cline{3-7}
                          & \multirow{2}{*}{InstructBLIP} & Vanilla           &$79.65_{(\pm0.24)}$ &$77.14_{(\pm0.43)}$ &$84.29_{(\pm0.36)}$ &$80.56_{(\pm0.18)}$  \\
                            &                               & \textbf{Ours}              &$\textbf{85.67}_{(\pm 0.28)}$ &$90.65_{(\pm 0.33)}$ &$79.53_{(\pm 0.19)}$ &$\textbf{84.73}_{(\pm 0.22)}$  \\ \cline{3-7}
                           & \multirow{2}{*}{mPLUG-Owl2} & Vanilla           &$80.43_{(\pm 0.26)}$ &$75.01_{(\pm 0.30)}$ &$91.26_{(\pm 0.15)}$ &$82.34_{(\pm 0.39)}$  \\
                           &                               & \textbf{Ours}              &$\textbf{89.00}_{(\pm 0.23)}$ &$90.18_{(\pm 0.41)}$ &$87.53_{(\pm 0.12)}$ &$\textbf{88.84}_{(\pm 0.45)}$  \\ 
\cline{1-7}
\multirow{8}{*}{\textit{Popular}}      & \multirow{2}{*}{LLaVA1.5}     & Vanilla           &$78.17_{(\pm0.17)}$ &$77.64_{(\pm0.26)}$ &$79.12_{(\pm0.35)}$ &$78.37_{(\pm0.18)}$  \\                                    
                           &                               & \textbf{Ours}              &$\textbf{79.43}_{(\pm 0.32)}$ &$72.94_{(\pm 0.25)}$ &$93.61_{(\pm 0.40)}$ &$\textbf{81.99}_{(\pm 0.07)}$  \\ \cline{3-7}
                          & \multirow{2}{*}{Qwen-VL}     & Vanilla           &$75.99_{(\pm0.33)}$ &$78.62_{(\pm0.41)}$ &$71.40_{(\pm0.38)}$ &$74.84_{(\pm0.34)}$  \\
                           &                               & \textbf{Ours}              &$\textbf{79.80}_{(\pm 0.29)}$ &$83.40_{(\pm 0.37)}$ &$74.40_{(\pm 0.12)}$ &$\textbf{78.65}_{(\pm 0.45)}$  \\ \cline{3-7}
                          & \multirow{2}{*}{InstructBLIP} & Vanilla           &$73.87_{(\pm0.58)}$ &$69.63_{(\pm0.54)}$ &$84.69_{(\pm0.68)}$ &$76.42_{(\pm0.52)}$  \\
                            &                               & \textbf{Ours}              &$\textbf{73.98}_{(\pm 0.26)}$ &$78.08_{(\pm 0.33)}$ &$78.08_{(\pm 0.18)}$ &$\textbf{78.08}_{(\pm 0.23)}$ \\  \cline{3-7}
                           & \multirow{2}{*}{mPLUG-Owl2} & Vanilla           &$71.96_{(\pm 0.20)}$ &$66.01_{(\pm 0.40)}$ &$90.53_{(\pm 0.15)}$ &$76.35_{(\pm 0.48)}$ \\
                           &                               & \textbf{Ours}              &$\textbf{77.87}_{(\pm 0.35)}$ &$73.46_{(\pm 0.47)}$ &$87.27_{(\pm 0.10)}$ &$\textbf{79.77}_{(\pm 0.22)}$  \\ 
\cline{1-7}
\multirow{8}{*}{\textit{Adversarial}}      & \multirow{2}{*}{LLaVA1.5}     & Vanilla          &$75.08_{(\pm0.33)}$ &$73.19_{(\pm0.49)}$ &$79.16_{(\pm0.35)}$ &$76.06_{(\pm0.24)}$    \\
                               &                               & \textbf{Ours}              &$\textbf{77.53}_{(\pm 0.21)}$ &$67.81_{(\pm 0.34)}$ &$93.13_{(\pm 0.43)}$ &$\textbf{78.48}_{(\pm 0.08)}$  \\ \cline{3-7}
                          & \multirow{2}{*}{Qwen-VL}     & Vanilla           &$75.46_{(\pm0.63)}$ &$77.92_{(\pm0.73)}$ &$71.07_{(\pm0.97)}$ &$74.33_{(\pm0.71)}$  \\
                           &                               & \textbf{Ours}              &$\textbf{79.20}_{(\pm 0.25)}$ &$82.30_{(\pm 0.36)}$ &$74.40_{(\pm 0.14)}$ &$\textbf{78.15}_{(\pm 0.47)}$  \\
                            \cline{3-7}
                          & \multirow{2}{*}{InstructBLIP} & Vanilla          &$70.56_{(\pm0.53)}$ &$66.12_{(\pm0.32)}$ &$84.33_{(\pm1.05)}$ &$74.12_{(\pm0.58)}$  \\
                            &                               & \textbf{Ours}              &$\textbf{74.71}_{(\pm 0.28)}$ &$70.37_{(\pm 0.32)}$ &$85.31_{(\pm 0.17)}$ &$\textbf{77.12}_{(\pm 0.23)}$ \\ \cline{3-7}
                           & \multirow{2}{*}{mPLUG-Owl2} & Vanilla           &$54.86_{(\pm 0.22)}$ &$53.36_{(\pm 0.39)}$ &$77.13_{(\pm 0.13)}$ &$63.08_{(\pm 0.41)}$ \\
                           &                               & \textbf{Ours}              &$\textbf{75.26}_{(\pm 0.30)}$ &$71.17_{(\pm 0.33)}$ &$84.93_{(\pm 0.16)}$ &$\textbf{77.44}_{(\pm 0.37)}$  \\ 
\hline
\end{tabular}
}
\caption{Results on GQA source of POPE benchmark. The best performances are \textbf{bolded}.}
\label{tab:pope_gqa}
\end{table*}

\end{document}